\documentclass[10pt,twocolumn,letterpaper]{article}

\usepackage[norule,symbol,perpage]{footmisc}
\usepackage{btas}
\usepackage{times}
\usepackage{epsfig}
\usepackage{graphicx}
\usepackage{amsmath}
\usepackage{amssymb}
\usepackage{subcaption}
\usepackage[skip=0.5\baselineskip]{caption}

\makeatletter
\def\ps@IEEEtitlepagestyle{
	\def\@oddfoot{\mycopyrightnotice}
	\def\@evenfoot{}
}
\def\mycopyrightnotice{
	{\hfill \footnotesize 978-1-7281-1522-1/19/\$31.00 \copyright 2019 IEEE\hfill}
}
\makeatother

\btasfinalcopy 

\ifbtasfinal\fi
\begin{document}

\title{Identity-Aware Deep Face Hallucination via Adversarial Face Verification}

\author{Hadi Kazemi\\
West Virginia University\\
{\tt\small hakazemi@mix.wvu.edu}
\and
Fariborz Taherkhani\\
West Virginia University\\
{\tt\small fariborztaherkhani@gmail.com}
\and
Nasser M. Nasrabadi \\
West Virginia University\\
{\tt\small nasser.nasrabadi@mail.wvu.edu}
}

\maketitle
\thispagestyle{empty}

\begin{abstract}
In this paper, we address the problem of face hallucination by proposing a novel multi-scale generative adversarial network (GAN) architecture optimized for face verification. First, we propose a multi-scale generator architecture for face hallucination with a high up-scaling ratio factor, which has multiple intermediate outputs at different resolutions. The intermediate outputs have the growing goal of synthesizing small to large images. Second, we incorporate a face verifier with the original GAN discriminator and propose a novel discriminator which learns to discriminate different identities while distinguishing fake generated HR face images from their ground truth images. In particular, the learned generator cares for not only the visual quality of hallucinated face images but also preserving the discriminative features in the hallucination process. In addition, to capture perceptually relevant differences we employ a perceptual similarity loss, instead of similarity in pixel space.
We perform a quantitative and qualitative evaluation of our framework on the LFW and CelebA datasets. The experimental results show the advantages of our proposed method against the state-of-the-art methods on the 8x downsampled testing dataset.
\end{abstract}
\let\thefootnote\relax\footnotetext{\mycopyrightnotice}
\section{Introduction}

Face hallucination methods deal with super-resolving a low-resolution (LR) face image and generating a high-resolution (HR) one. They have many applications such as face recognition, face tracking, security in surveillance video, and facial expression estimation. 
One of the most common issues in the practical face recognition systems is that they have low performance on low-resolution face images captured in the wild. Especially in the standard surveillance videos, detected faces might have a resolution of 20$\times$20 pixels or smaller \cite{zhou2015learning}. Such LR face images negatively affect the performance of the subsequent face recognition and analysis. Consequently, in the past few years generating HR face images from LR ones has attracted great research interests.

Traditional interpolation techniques, such as the nearest neighbor or bilinear up-scaling, are not able to reconstruct high-frequency details. On the contrary, frameworks which are based on example-based super-resolution (SR) schemes \cite{baker2000limits} have shown a good performance in fine detailed reconstruction from a LR image compared to the interpolation-based methods. This capacity is acquired by learning the patterns, textures, and geometrical characteristics of face images based on different machine learning techniques trained on a comprehensive pair of training HR/LR images.

Unlike the natural images, SR face hallucination images have similar structures. Employing only a reconstruction error may result in faces with visually undesirable artifacts. For example, small geometry distortion in face components which plays a critical role in person identification, such as the mouth and eyes, can degrade the subjective quality of the face hallucination. Therefore, the global face shape and local characteristics such as textures and local geometric structures (e.g., nose and eyes) need to be handled cautiously in face hallucination \cite{baker2000hallucinating, wang2014comprehensive}.

Surveillance cameras, which usually provide low-resolution images, especially for small objects of interest such as faces taken at a distance, makes face identification a more challenging problem. This is due to the lack of sufficiently discriminative features in low-resolution face images. An empirical study \cite{zou2012very} showed that for effective face identification, the minimum face resolution should be between 32 x 32 and 64 x 64 pixels. Thus, a lower resolution face will significantly degrade the recognition performance for the current recognition models. Consequently, an effective face hallucination framework is desirable.

Typically, for the high upscaling factors, the textural detail in the reconstructed SR images is absent. The restricted use of mean squared error (MSE) between the generated HR image and the ground truth as the only optimization target could be the reason for the missing detailed information in the reconstructed SR images. More specifically, the MSE lacks the ability to capture perceptually relevant differences, e.g., textural detail, as it is defined at image pixel level \cite{wang2004image, wang2003multiscale}. 

Majority of existing face hallucination techniques \cite{wang2014comprehensive} have been focused on hallucinating faces which are visually pleasant. In other words, they just generate HR details neglecting whether the added details are useful for face recognition. Such reconstructed faces, usually do not improve the face recognition/verification performance. On the contrary, incorporating the identity in face hallucination process enable the framework to preserve the facial
details which play a crucial role in face recognition and serve the purpose much better. Therefore, for many real-world applications, preserving identity in face reconstruction is a vital step of hallucination process \cite{hennings2008simultaneous, wu2016deep}.

\begin{figure*}[t]
	\begin{center}
		
		\includegraphics[width=0.99\linewidth]{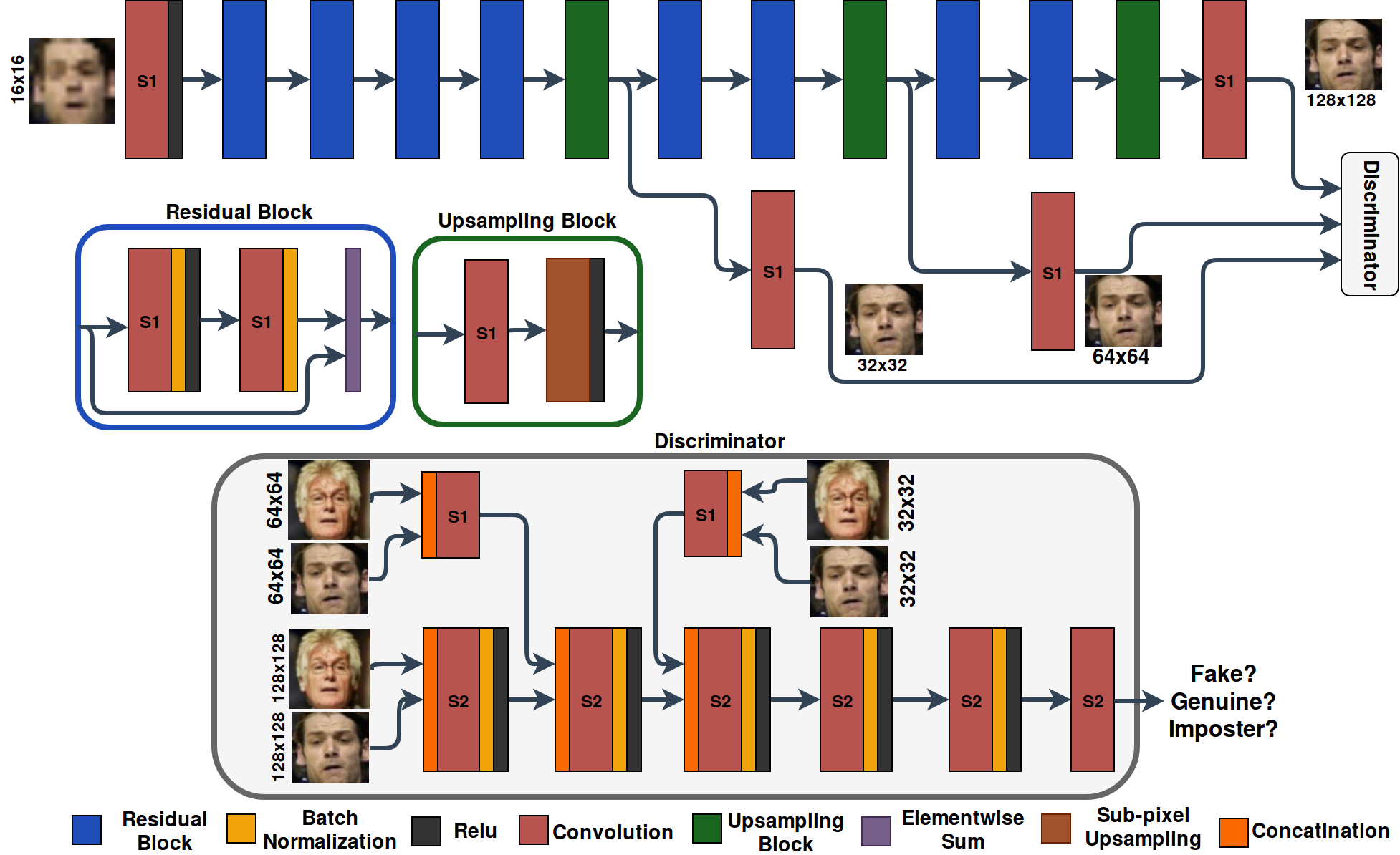}
	\end{center}
	\caption{Architecture of the proposed network. Different branches of the generator produce images of different scales. The discriminator, then, learns to distinguish fake and real images, jointly in multiple scales.}
	\label{fig:generator}
\end{figure*}

In this work, first, we propose a multi-scale generative adversarial network architecture, for face hallucination with a high up-scaling ratio factor. The generator network has multiple outputs, that share most of their parameters, in
a progressive structure. As shown in Fig. \ref{fig:generator}, the input to the network is a low-resolution face image, and multiple face images with different scaling factors are generated through different branches of the network. The deepest output of our generator has resolution equal to our high-resolution face image. The intermediate outputs, have a growing goal of synthesizing small to large images.

Second, we incorporate a face verifier with the original GAN discriminator and propose a novel discriminator which learns to discriminate betwen different identities while distinguishing fake generated HR face images from their corresponding ground truths. Correspondingly, the generator is trained to not only generate face images of high visual quality but also preserve the discriminative features in the hallucination process. Intuitively, improving the discriminator enhances the verification ability by infusing missing details to the LR image, and improving the verification performance boosts the discriminator (which trains the generator) to look for the quality of identity discriminative features in the generated images.
In summary, our framework has three major contributions:
\begin{itemize}
	\item We propose a novel identity-aware GAN for face super resolution which enable us to hallucinate photo-realistic HR faces while preserving the face identity.
	\item We combine the disciminator and face verifier by proposing a single network which performs both tasks simultaneously.
	\item Our discriminator jointly learns to distinguish face images at multiple scales. This unified multi-scale structure enables the discriminator to transfer information between generated face images of different scales.
	\item A series of qualitative and quantitative experiments proves the
	effectiveness of the proposed end-to-end framework.
\end{itemize}

\subsection{Related Work}

Prediction-based methods were among the early methods for single image super-resolution. However, these filtering approaches, oversimplify this problem and usually yield outputs with low details and blurry textures. Some other frameworks have been proposed in \cite{allebach1996edge, li2001new} that focus particularly on edge-preservation.
More effective approaches, which are usually data-driven, learn a complex mapping between low- to high-resolution images. Early approaches to the SR problem were developed based on compressed sensing \cite{yang2008image, zeyde2010single, dong2011image}. Huang et al. \cite{huang2015single} exploit self-similarity, where self dictionaries are extended for small transformations and shape variations. A method based on convolutional sparse coding is proposed in \cite{gu2015convolutional} to improve output consistency by processing the whole image at once instead of overlapping patches.

Deep learning-based approaches outperformed most of the traditional methods in computer vision \cite{talreja2017multibiometric, taherkhani2018deep, zohriadeh2019Class, talreja2018biometrics, taherkhani2017restoring,  talreja2018using, taherkhani2018facial, taherkhani2019matrix}, more specifically face hallucination schemes \cite{wang2014comprehensive, wu2016deep, tuzel2016global, yu2016ultra, zhu2016deep}. In \cite{wu2016deep}, a deep joint face hallucination and recognition scheme is proposed, which comprises two separate networks, namely SR and face recognition networks. They have jointly optimized the two networks iteratively, however, due to employing a relatively shallow CNN, it resulted in unsatisfactory visual quality in face reconstruction. A much deeper CNN is utilized in \cite{zhu2016deep} to generate HR face image of higher visual quality. To this end, they trained a cascaded bi-network progressively to learn a dense correspondence during the training phase. Song et al. \cite{song2017learning} proposed a two-stage face hallucination process that first reconstructs facial parts employing a deep CNN, and then refines the reconstructed faces using a fine-grained facial structure learner.

Recently, generative adversarial network (GAN) has been successfully adopted by many computer vision applications such as image synthesis, image SR, and in-painting \cite{goodfellow2014generative, kazemi2018unsupervised, kazemi2019style}. The SR-GAN \cite{ledig2017photo} is the pioneer in utilizing GAN in inferring photo-realistic high-resolution natural images from LR images. They incorporated the perceptual loss in addition to the adversarial loss to push the solution toward learning to preserve the content of images in the super-resolving process. However, Yu et al. \cite{yu2016ultra} showed that this framework is not effective for super-resolving LR to face images. They introduced a pixel-wise $L_2$ regularization term to exploit the discriminator network feedback and produce faces with higher similarity to the real ones.

Similarly, in \cite{tuzel2016global}, deconvolutional layers are separately applied to super-resolve local and global parts. However, none of the mentioned methods guarantee identity preservation in the reconstruction process. Moreover, they often generate unrealistic low quality face images from the very low resolution face images, as much of the facial structural information is missing.

An end-to-end GAN-based SR model combined with a face alignment network is proposed in \cite{bulat2018super} that employs a heatmap loss to integrate facial geometrical information in hallucination process by detecting facial landmarks. 
The application of deep reinforcement learning in HR face generation has also been investigated in \cite{cao2017attention}. They proposed to employ a recurrent policy network for individual HR face regions reconstruction based on previous regions reconstructions. Finally, they applied a local enhancement network to improve the facial details. Again, the importance of identity preservation has been neglected in these works.

\section{Preliminaries}
In this section, we provide some rudiments of GANs, necessary to understand the proposed preference-based image generation framework.
\subsection{Generative Adversarial Networks (GANs)}\label{sec:GAN}
GANs \cite{goodfellow2014generative} are a type of generative models which learn the statistical distribution of the training data, allowing us to synthesize data samples by mapping a random noise $z$ to an output image $y$: $G(z): z \longrightarrow y$, where $G$ is the generator network. GAN in its conditional setting (cGAN) is proposed in \cite{isola2017image} which learns a mapping from an input $x$ and a random noise $z$ to the output image $y$: $G(x, z): \{x, z\} \longrightarrow y$, using an autoencoder network. The generator model $G(x, z)$, is trained to generate images which are not distinguishable from the \textit{real} samples by a discriminator network, $D$. Simultaneously, the discriminator is learning, adversarially, to discriminate between the \textit{fake} generated images by the generator and the real samples from the training dataset. The objective function of GAN is given by:
\begin{align} \label{eq:ad_loss}
l_{GAN}(G,D) &=  \mathbf{E}_{x,y\sim p_{data}}[\log D(x,y)] \\ \nonumber &+  \mathbf{E}_{x, z\sim p_{z}}[\log (1 - D(x,G(x, z)))],
\end{align}
where $G$ attempts to minimize it and $D$ tries to maximize it. Since the adversarial loss is not enough to guarantee that the trained network generates the desired output, one may add an extra Euclidean distance term to the objective function to generate images which are near the ground truth. Consequently, the final objective is defined as follows:
\begin{align} \label{eq:cgan1}
G^* = \arg \min_G \max_D l_{GAN}(G,D) + \lambda l_{L1}(G),
\end{align}
where $l_{L1}(G) = \parallel y - G(x,z) \parallel_1$ and $\lambda$ is weighting factor.

\section{Proposed Multi-Scale GAN Architecture}
The goal of this work is to learn a generating function $G$ to reconstruct a HR face images from a given LR input face image. The backbone of our deep generator network $G$, which is demonstrated in Figure \ref{fig:generator}, is a series of residual blocks with an identical layout. The residual blocks comprise of two $3\times3$ convolutional layers followed by batch-normalization layers \cite{ioffe2015batch} and ParametricReLU activation function \cite{he2015delving}. We employ three pre-trained sub-pixel convolution layers \cite{shi2016real} to gradually increase the resolution of the input image. 
 
The preliminary results showed that the network is not able to generate high-quality images for upscaling factor of greater than 4x. Consequently, we propose to progressively learn a series of multi-scale images.  As shown in Figure \ref{fig:generator}, our generator has multiple outputs at different resolutions. Each output of the generator learns the face image distribution at that scale. 
We also concatenate the images at different depths of the discriminator. This multi-scale structure improves the discriminator by jointly learning to distinguish face images at multiple scales. This enable us to transfer information between images of different scales. 

The architecture of our discriminator is shown in Figure \ref{fig:generator}. Following the previous works in \cite{radford2015unsupervised, ledig2017photo}, we use LeakyReLU activation  ($\alpha= 0.2$) and avoid max-pooling in the architecture. However, our discriminator takes images of different resolutions as its inputs. To this end, we utilize strided convolutions ($s=2$) in the main branch of the discriminator that reduce the spatial size of the feature maps by a factor of two. Simultaneously, images of lower resolutions are processed by convolutional layers ($s=1$) which extract feature maps of the same size. We then concatenate the extracted feature maps of the lower resolution images with the feature maps of the same spatial size in the main branch.

\section{Training Loss Function}
To train the proposed network, we utilize multiple loss terms, including an adversarial face verification loss, perceptual loss, and color-consistency regularization.
\subsection{Perceptual Loss}
The Pixel-wise MSE loss is one of the most widely used loss terms for image super-resolution problems. However,  despite the high PSNR, the learned solutions by MSE optimization often lack the high-frequency information, which results in unsatisfactory images of excessively smooth textures. Consequently, similar to \cite{ledig2017photo}, rather than relying on the pixel-level losses, we use MSE on the high level extracted features via a pre-trained VGG19 \cite{simonyan2014very}, denoted by $\Phi$, which represents perceptual similarity between the generated HR image and its corresponding ground truth. Let $\Phi_j(x)$ denote the feature maps of the $j^{th}$ layer of the loss network for the input image $x$. The perceptual loss, which has been introduced in \cite{gatys2015texture}, is defined as the Euclidean distance between the feature representations of a super-resolved image $G(I_{LR})$ and the reference image $I_{HR}$:
\begin{align} \label{eq:perc_loss} \nonumber
l_{p}^j(G(I_{LR}), I_{HR}) = \frac{1}{N_j}\parallel \Phi_j(G(I_{LR})) - \Phi_j(I_{HR}) \parallel_2^2,
\end{align}
where $N_j$ is the number of perceptrons in the $j$th layer.

\subsection{Adversarial Face Verification Loss}
In addition to the perceptual loss described above, one may want to add the adversarial loss of GAN, described by Equation \ref{eq:ad_loss}, to train the generator network.  This loss encourages the generator to favor, by seeking to fool the discriminator, production of images which reside on the manifold of natural face images. 

However, in this work, we introduce a completely different loss function, namely adversarial face verification loss (AFVL). Our proposed AFVL follows two different goals simultaneously. First, it should help the generator to learn high-frequency information, which cannot be learned by the sole adoption of MSE or perceptual loss. In this way, the generator would be able to produce highly realistic HR images. However, forcing the generator to just care about visual realism may come with the cost of losing identity information. Moreover, incorporating identity information in the discriminator decisions results in training a generator which preserves the critical facial features, which matters in person identification, in super resolving process.

Our discriminator, instead of assigning a \textit{fake} or \textit{real} label to the patches of its HR input image, performs a three ways classification task. More specifically, it classifies the HR images into \textit{fake}, \textit{genuine}, and \textit{imposter}. Particularly, the discriminator takes a pair of images instead of a single image. To train the discriminator, we define four different pairs as follows:
\begin{align}
	&p_1 = (I_{HR}^{i1}, G(I_{LR}^{i1})) \qquad p_2 = (I_{HR}^{i}, G(I_{LR}^{j})) \\ \nonumber 
	&p_3 = (I_{HR}^{i1}, I_{HR}^{i2}) \qquad \quad p_4 = (I_{HR}^{i}, I_{HR}^{j}),
\end{align}
where $I_{HR}^i$ represents a high-resolution face image of the $i^{th}$ identity in our dataset, and the number next to the $i$ shows if the same image is used in both side of the pair or not. More specifically, the first pair, $p_1$, includes a HR face image of the $i^{th}$ identity and the super-resolved image from the down-sampled version of the exact same image. On the contrary, the second pair $p_2$ includes a HR face image of the $i^{th}$ identity and the super-resolved image from the down-sampled version of an image of another identity. The discriminator is supposed to classify these two pairs as a \textit{fake} samples. These pairs aim to help the discriminator to learn the visual realism of the generated images. 

The third pair, $p_3$, comprises two different HR face images of the same identity. Note that none of the images are generated by the generator network. The discriminator is trained to classify this pair as \textit{genuine}. On the contrary, the last pair $p_4$ does not share the identity between the HR images. Clearly, it is desired that the discriminator classify $p_4$ as  \textit{imposter}. Hence, the pairs $p_3$ and $p_4$ jointly enable the discriminator to capture critical features which are highly influential in face verification task.

To train the discriminator using these four pairs, we can
define the adversarial face verification loss as:

\begin{align} \label{eq:afvl_dis}
&l_{AFVL}^d(G,D) =  \mathbf{E}_{(x, y)\sim p_1}[\log d_{f}(x,y)] \\ \nonumber &+  \mathbf{E}_{(x, y)\sim p_2}[\log d_{f}(x,y)]
+  \mathbf{E}_{(x, y)\sim p_3}[\log d_{gen}(x,y)] \\ \nonumber &+  \mathbf{E}_{(x, y)\sim p_4}[\log d_{imp}(x,y)],
\end{align}
where $d_{f}$, $d_{gen}$, and $d_{imp}$ are the outputs of discriminator for \textit{fake}, \textit{genuine}, and \textit{imposter} classes, respectively. Similar to the original GAN, the discriminator is trained to maximize this objective function. However, to train the generator, we only use the first two pairs $p_1$ and $p_2$. Similar to the training process of the discriminator, here, we train a generator which consider the identity-preservation in its face hallucination process. To this end, the generator is trained to maximize $d_{gen}$ for $p_1$, and $d_{imp}$ for $p_2$. In this way, the generator not only tries to fool the discriminator in terms of visual realism of the generated images, by minimizing $d_{f}$, but also takes into account the identity of the super-resolved image, through maximization of the verification objective function. In short, the generator maximize the following objective function:
\begin{align} \label{eq:afvl_gen}
&l_{AFVL}^g(G,D) =  \mathbf{E}_{(x, y)\sim p_1}[\log d_{gen}(x,y)] \\ \nonumber &+  \mathbf{E}_{(x, y)\sim p_2}[\log d_{imp}(x,y)].
\end{align}

\subsection{Color-consistency regularization}
As  we go deeper into our generator, the resolution of the generated image is also gradually increased. Since all the  generated  images belong to the same input but at different scales, they  require to  have  similar structures and colors. To this end, we utilize color-consistency regularization term  as  an additional objective function to  keep the generated  samples of different scales from the same input to be more consistent in color. This can improve the quality of the generated images.

Let $\mu = \sum_k x_k/N$, and $\Sigma = \sum_k (x_k - \mu)(x_k - mu)^T/N$ represent the  mean  and  covariance  of  pixels  of  the  given image, respectively, where $x_k = (R, G, B)^T$ is a pixel in the generated image. Then, the  color-consistency  regularization  term tries to minimize the differences of $\mu$ and $\Sigma$ between the different generated images at various resolutions, which inspires the consistency:
\begin{align} \label{eq:color_loss}
l_{C_i} = \dfrac{1}{n} \sum_{j=1}^{n} &\Big( \lambda_1 \parallel \mu_{s_i^j} - \mu_{s_{i-1}^j}\parallel_2^2 \\ \nonumber&+ \lambda_2 \parallel \Sigma_{s_i^j} - \Sigma_{s_{i-1}^j}\parallel_F^2\Big),
\end{align}
where $\mu_{s_i^j}$ and $\Sigma_{s_i^j}$ represents the mean and covariance matrix of the $j^{th}$ sample generated in $i^{th}$ scale, and, $n$ is the batch size. Note that images of different resolution are generated by different branches of the generator. In our work, since the generator produces images at three scales, we have two color-consistency regularization terms corresponding to $i=1, 2$, where each $i$ belongs to the image of size $2^{i+6}$ pixels.

\subsection{Total Loss}
While the discriminator is trained using only the AFVL loss $l_{AFVL}^d$, the total loss to train the generator is defined as: 
\begin{align} \label{eq:loss_tot}
l_t^g = \sum_j l_{p}^j + \lambda_c \sum_i l_{C_i} - \lambda_a l_{AFVL}^g(G,D).
\end{align}
Note that we minimize the perceptual loss in more than one layer to enforce fine and coarse perceptual similarity.
\section{Experiments}
Our experiments aim to show that our framework can generate high visual quality HR faces at different up-scaling factors while preserving the identity of the hallucinated faces. We compare our method against baselines both qualitatively and quantitatively.
\subsection{Datasets}
Our experiments are evaluated on the Labeled Faces in the Wild (LFW funneled) \cite{learned2016labeled} and the BioID \cite{jesorsky2001robust} datasets.
The LFW dataset contains 13,233 face images which are collected from the web. Images in this dataset cover a vast variety of pose variations and facial expressions. This dataset comprises of four different parts, including the original set and three different aligned images. In this work, we only use the original ones to conduct our experiments. To generate the LR and HR pairs, we use the original aligned images of size $250 \times 250$ pixels, and extract the centric $128\times 128$ image patches as the HR images. Then, we create the corresponding LR images by down-sampling the HR ones using a bilinear kernel with the down-sampling factor. Our training set includes 9,526 images which leaves us the remaining 3,707 images for testing.

The BioID dataset consists of 1,521 face images. We use 1,028 images for training and the remaining 493 images for testing. This follows the same split provided by the LFW dataset. The images are aligned with SDM method \cite{xiong2013supervised} and then a patch with the size of $160\times 120$ is cropped from the center of each image.

\subsection{Visual Realism Evaluation}
For visual realism evaluation, we evaluate our proposed SR network on two scaling factors of $4x$ and $8x$. Input low-resolution image is generated by resizing the original images with the scaling factors. Hence, to generate LR images, the HR images of BioID are resized to $40\times 30$ and $20\times 15$, and the HR images are resized to $32\times 32$ and $16\times 16$, respectively.

For the evaluation metrics, we adopt the widely used Peak Signal-to-Noise Ratio
(PSNR), structural similarity (SSIM) as well as feature similarity (FSIM) \cite{zhang2011fsim}.
We perform a comparison between our method and several state-of-the-art face hallucination and image super-resolution techniques. Particularly, we compare with the BCCNN \cite{zhou2015learning}, SFH \cite{yang2013structured},
GLN \cite{tuzel2016global}, MZQ \cite{ma2010hallucinating} face hallucination approaches and three general image super-resolution methods: A-FH \cite{cao2017attention}, SRCNN \cite{dong2014learning}, and VDSR \cite{kim2016accurate}.

\begin{figure*}
	\begin{subfigure}{.99\textwidth}
		\centering
		\includegraphics[width=.9\linewidth]{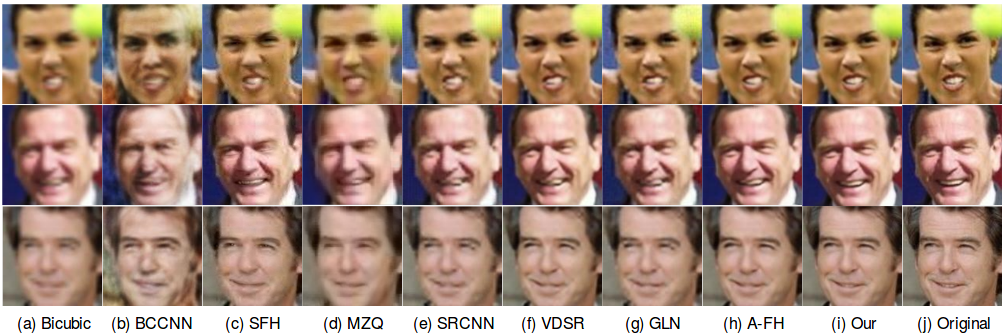}
	\end{subfigure}%
	\\
	\begin{subfigure}{.99\textwidth}
		\centering
		\includegraphics[width=.9\linewidth]{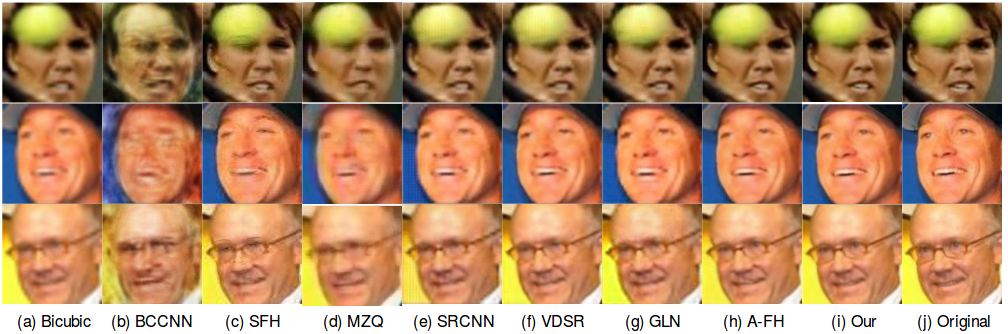}
	\end{subfigure}
	\caption{Qualitative results on LFW-funneled with scaling factor of 4.}
	\label{fig:lfw4}
\end{figure*}
\begin{figure*}
	\begin{center}
		\includegraphics[width=0.9\linewidth]{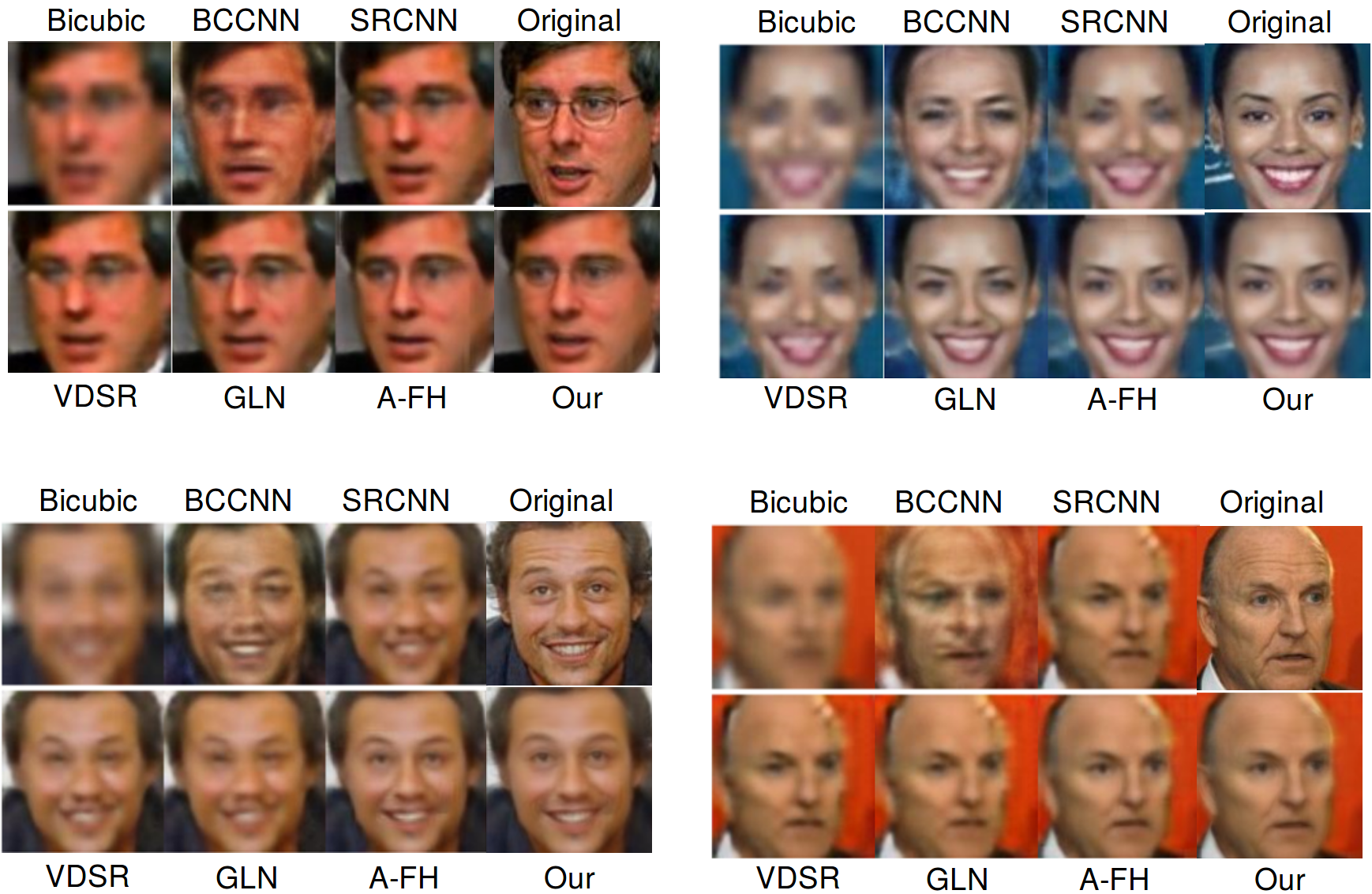}
	\end{center}
	\caption{Qualitative results on LFW-funneled with scaling factor of 8.} 
	\label{fig:lfw8}
\end{figure*}

\begin{table*}[]
	\def\arraystretch{1.3}
	\caption{Comparison between our method and others in terms of PSNR, SSIM and FSIM evaluation metrics.}
	\begin{center}
		\setlength\tabcolsep{5.5pt}
		\begin{tabular}{|l|c|c|c|c|c|c|c|c|c|c|c|c|}
			\hline
			Method       & \multicolumn{3}{c|}{LFW-funneled 4x} & \multicolumn{3}{c|}{LFW-funneled 8x} & \multicolumn{3}{c|}{BioID 4x} & \multicolumn{3}{c|}{BioID 8x} \\ \hline
			& PSNR       & SSIM       & FSIM       & PSNR       & SSIM       & FSIM       & PSNR    & SSIM     & FSIM     & PSNR    & SSIM     & FSIM     \\ \hline
			Bicubic      & 26.79      & 0.8469     & 0.8947     & 21.92      & 0.6712     & 0.7824     & 25.18   & 0.8170   & 0.8608   & 20.68   & 0.6434   & 0.7539   \\ \hline
			SFH \cite{yang2013structured}         & 26.59      & 0.8332     & 0.8917     & 22.12      & 0.6732     & 0.7832     & 25.41   & 0.8034   & 0.8494   & 20.31   & 0.6234   & 0.7238   \\ \hline
			BCCNN \cite{zhou2015learning}       & 26.60      & 0.8329     & 0.8982     & 22.62      & 0.6801     & 0.7903     & 24.77   & 0.8034   & 0.8421   & 21.40   & 0.6504   & 0.7621   \\ \hline
			MZQ \cite{ma2010hallucinating}         & 25.93      & 0.8313     & 0.8865     & 22.12      & 0.6771     & 0.7802     & 24.66   & 0.8001   & 0.8573   & 21.11   & 0.6481   & 0.7594   \\ \hline
			SRCNN \cite{dong2014learning}       & 28.94      & 0.6363     & 0.9069     & 23.92      & 0.6927     & 0.8314     & 27.02   & 0.8517   & 0.8771   & 22.34   & 0.6980   & 0.8274   \\ \hline
			VDSR \cite{kim2016accurate}        & 29.25      & 0.8711     & 0.9123     & 24.12      & 0.7031     & 0.8391     & 28.52   & 0.8627   & 0.8914   & 24.31   & 0.7321   & 0.8465   \\ \hline
			GLN \cite{tuzel2016global}         & 30.34      & 0.8922     & 0.9151     & 24.51      & 0.7109     & 0.8405     & 29.13   & 0.8794   & 0.8966   & 24.76   & 0.7421   & 0.8525   \\ \hline
			A-FH \cite{cao2017attention} & 32.93      & 0.9104     & 0.9427     & 26.17      & 0.7604     & 0.8630     & 31.56   & 0.9002   & 0.9343   & 26.56   & 0.7864   & 0.8747   \\ \hline
			Our          & \textbf{33.59}      & \textbf{0.9213}     & \textbf{0.9601}     & \textbf{26.94}      & \textbf{0.7723}     & \textbf{0.8772}     & \textbf{32.49}   & \textbf{0.9899}   & \textbf{0.9481}   & \textbf{27.83}   & \textbf{0.7967}   & \textbf{0.8914}   \\ \hline
		\end{tabular}
	\end{center}
	\label{tbl:eval}
\end{table*}

Table \ref{tbl:eval} compares the performance of our method with other state-of-the-art techniques. Our proposed framework significantly outperforms all the other methods in terms of PSNR, SSIM and FSIM metrics on LFW and BioID datasets.
Since the traditional face hallucination methods, i.e., SFH and
MZQ, are highly dependent on the facial
landmarks detection performance, and the landmarks detection is not quite reliable in very low-resolution images, their performances on 8x up-scaling factor are too low compared to the other methods. Among deep-learning based methods, our work outperforms the current state-of-the-art image super-resolution method (A-FH) on different experiments.

In addition, our method significantly outperforms state-of-the-art face hallucination methods, namely SiGAN and GLN. Figures \ref{fig:lfw4} and \ref{fig:lfw8} illustrates the qualitative comparisons of face hallucination results on the LFW dataset for 4x and 8x up-scaling factors, respectively. Our proposed framework generates face images that are more clear and sharper compared to the A-FH, GLN, and VDSR. 

\subsection{Identity Preserving Evaluation}

To evaluate the performance of different methods in preserving the identity of the LR face in the hallucination process, we compare our method with several state-of-the-art face hallucination methods including DFCG \cite{song2017learning}, SiGAN \cite{hsu2018sigan}, UR-DGN \cite{yu2016ultra}, GLN \cite{tuzel2016global}, A-FH \cite{cao2017attention}, DCGAN \cite{radford2015unsupervised}, PRSR \cite{dahl2017pixel}, and \cite{ledig2017photo}. Note that we selected the methods whose performances are reported in the literature. In addition, we evaluated the performances of the top two methods with the highest visual realism scores in the evaluation section, namely A-FH and GLN.

\subsubsection{Face Verification Performance}

To compare the performance of the proposed method with the previous face hallucination techniques for face verification task, we employ a state-of-the-art CNN-based face recognition engine, the OpenFaces \cite{amos2016openface}. We report the face recognition rate and verification rate of the hallucinated faces by different methods. The accuracy of the hallucinated HR faces is evaluated, following the standard face verification methodology described in \cite{amos2016openface}. The accuracy is calculated which is based on if the OpenFaces verifies the hallucinated HR faces as the same identity as their corresponding ground-truth or not.

We setup our experiment by first randomly sampling 200,000 face pairs from LFW, similar to the training set of the OpenFaces recognition engine as described in \cite{amos2016openface}. Then, 6,000 faces are randomly sampled from
the remaining face images of LFW for the face verification performance evaluation.
Our evaluation metric is the area under curve (AUC) \cite{learned2016labeled} of the trained face verification system based on the super-resolved HR faces.

Table \ref{tbl:auc} reports the AUCs for the generated HR faces of $128 \times 128$ and $64 \times 64$ pixels from $16 \times 16$ LR faces using different face hallucination techniques. The results show that the AUC for the generated HR faces by our method is significantly higher than the AUCs of the other methods. This proves the superiority of our method in preserving the identity of faces in the hallucination process

\begin{table*}[]
	\caption{Comparison of LFW face recognition rates for the hallucinated HR faces using different techniques}
	\begin{center}
	\def\arraystretch{1.2}
	\begin{tabular}{|l|c|c|c|c|c|c|c|c|}
		\hline
		           & \multicolumn{2}{c|}{AUC} & \multicolumn{3}{c|}{Verification Acc. 4x} & \multicolumn{3}{c|}{Verification Acc. 8x} \\ \hline
		Method     & 4x     & 8x     & Top-1  & Top-5  & Top-10 & Top-1  & Top-5  & Top-10 \\ \hline
		HR         & 98.8\% & 99.1\% & 36.8\% & 55.9\% & 63.8\% & 37.5\% & 57.0\% & 66.2\% \\ \hline
		Bicubic    & 75.7\% & 76.0\% & 11.6\% & 27.5\% & 37.6\% & 11.7\% & 27.1\% & 36.4\% \\ \hline
		DFCG \cite{song2017learning}       & 73.9\% & - 	 & 9.6\%  & 23.7\% & 34.8\% & - 	 & - 	  & - \\ \hline
		UR-DGN \cite{yu2016ultra}    & 72.8\% & - 	 & 12.2\% & 29.0\% & 38.7\% & - 	 & - 	  & - \\ \hline
		DCGAN \cite{radford2015unsupervised}     & 74.8\% & - 	 & 9.3\%  & 24.9\% & 33.9\% & - 	 & - 	  & - \\ \hline
		PRSR \cite{dahl2017pixel}      & 76.9\% & - 	 & 13.3\% & 29.7\% & 40.1\% & - 	 & - 	  & - \\ \hline
		SiGAN \cite{hsu2018sigan}     & 83.4\% & - 	 & 17.9\% & 32.9\% & 48.1\% & - 	 & - 	  & - \\ \hline
		GLN \cite{tuzel2016global}       & 80.6\% & 79.2\% & 17.5\% & 30.6\% & 45.7\% & 17.1\% & 29.9\% & 44.3\% \\ \hline
		SR-GAN \cite{ledig2017photo}    & 81.8\% & 71.8\% & 18.3\% & 31.4\% & 47.6\% & 15.8\% & 26.7\% & 39.6\% \\ \hline
		A-FH \cite{cao2017attention}      & 85.2\% & 85.6\% & 18.8\% & 31.9\% & 48.4\% & 18.9\% & 32.4\% & 48.9\% \\ \hline
		Our        & \textbf{86.0}\% & \textbf{86.5}\% & \textbf{19.5}\% & \textbf{33.2}\% & \textbf{49.5}\% & \textbf{20.1}\% & \textbf{33.5}\% & \textbf{50.1}\% \\ \hline
	\end{tabular}
	\end{center}
\label{tbl:auc}
\end{table*}

\subsubsection{Face Recognition Performance}
We also evaluate the performance of our framework for the face recognition task. We setup this experiment as suggested in \cite{amos2016openface}. For this experiment, the training set includes 11,000 face image of 680 different identities randomly sampled from the LFW dataset. The remaining 2,000 face images form our test set. The OpenFaces is trained on face images after resizing to $96 \times 96$. Likewise, all the hallucinated HR faces need to be resized to $96\times96$ at the test time to evaluate the performance.

Table \ref{tbl:auc} compares the top-1, top-5, and top-10 face recognition rates of different methods for $64 \times 64$, i.e., 4x up-scaling factor, and $128 \times 128$, 8x up-scaling factor, HR faces upscaled from $16 \times 16$ LR faces. Note that for some methods the results for 8x upscaling factors are not reported in the original papers. The result prove the superiority of our method, in terms of the average recognition rates for the hallucinated HR faces, compared to the previous state-of-the-art methods. Note that, GLN and SR-GAN have worse performance on 8x compared to the 4x upscaling factor due to the generated artifacts by these methods at 8x super-resolved face images.

Compared to the bicubic interpolation, DFCG and DCGAN have lower face recognition rates, which shows the importance of reconstructing facial features that are critical in face re-identification. In other word, despite the higher level of HR details in these methods, compared to the bicubic, they are not useful for identity recognition. 

\section{Conclusion}
In this paper, we have proposed a identity-preserving face hallucination GAN-based framework. We enabled our generator to up-scale LR face images by a factor of 8 and learn to jointly generate face images of progressive resolution. We have also proposed a new discriminator which can jointly learn to verify the identity of the generated images and check their visual quality. The new discriminator architecture enables the whole face hallucination process to be identity-preserving too. Experimental results on several LR version of face benchmarks have convincingly demonstrated the effectiveness of the proposed approach.
{\small
\bibliographystyle{ieee}
\bibliography{submission_example}
}

\end{document}